\documentclass{article}

\usepackage{arxiv}
\usepackage{amsmath}
\usepackage{amssymb}
\usepackage[utf8]{inputenc} % allow utf-8 input
\usepackage[T1]{fontenc}    % use 8-bit T1 fonts
\usepackage{hyperref}       % hyperlinks
\usepackage{url}            % simple URL typesetting
\usepackage{booktabs}       % professional-quality tables
\usepackage{amsfonts}       % blackboard math symbols
\usepackage{nicefrac}       % compact symbols for 1/2, etc.
\usepackage{microtype}      % microtypography
\usepackage{lipsum}
\usepackage{graphicx}
\graphicspath{ {./images/} }

\title{A Generative Neural Network Approach for 3D Multi-Criteria Design Generation and Optimization of an Engine Mount for an Unmanned Air Vehicle}

\author{
  Christoph Petroll \\
  Institute of Lightweight Engineering\\
  University of the Bundeswehr Munich\\
  München, BY 85579 \\
  \texttt{christophpetroll@bundeswehr.org} \\
  \And
  Sebastian Eilermann \\
  Computer Science in Mechanical Engineering\\
  Helmut-Schmidt University\\
  Hamburg, HH 22043  \\
  \texttt{sebastian.eilermann@hsu-hh.de} \\
  \And
  Philipp Hoefer \\
  Institute of Lightweight Engineering\\
  University of the Bundeswehr Munich\\
  München, BY 85579 \\
  \texttt{philipp.hoefer@unibw.de} \\
  \AND
   Oliver Niggemann \\
   Computer Science in Mechanical Engineering\\
   Helmut-Schmidt University \\
   Hamburg, HH 22043 \\
   \texttt{oliver.niggemann@hsu-hh.de} \\
}

\begin{document}

\maketitle

\def\thefootnote{*}\footnotetext{These authors contributed equally to this work}\def\thefootnote{\arabic{footnote}}

\begin{abstract}
One of the most promising developments in computer vision in recent years is the use of generative neural networks for functionality condition-based 3D design reconstruction and generation. Here, neural networks learn dependencies between functionalities and a geometry in a very effective way. For a neural network the functionalities are translated in conditions to a certain geometry. But the more conditions the design generation needs to reflect, the more difficult it is to learn clear dependencies. This leads to a multi criteria design problem due various conditions, which are not considered in the neural network structure so far.
In this paper, we address this multi-criteria challenge for a 3D design use case related to an unmanned aerial vehicle (UAV) motor mount. We generate 10,000 abstract 3D designs and subject them all to simulations for three physical disciplines: mechanics, thermodynamics, and aerodynamics. Then, we train a Conditional Variational Autoencoder (CVAE) using the geometry and corresponding multicriteria functional constraints as input. We use our trained CVAE as well as the Marching cubes algorithm to generate meshes for simulation based evaluation. The results are then evaluated with the generated UAV designs. Subsequently, we demonstrate the ability to generate optimized designs under self-defined functionality conditions using the trained neural network.
\end{abstract}

% keywords can be removed
%\keywords{First keyword \and Second keyword \and More}
\newpage
\section{Introduction}
\label{sec:intro}

The potential of using neural networks (NN) for computer aided design (CAD) generation shows new possibilities in the fields e.g. medicine, engineering as well as product development. Algorithms iteratively generate a variety of solutions in the shortest possible time for higher-performance designs \cite{Jang_2022}. Only functionality requirements with some boundary conditions are needed. This is achieved by NN as they connect the functionality requirements, as conditions, directly to generated geometry features. The weights of the neural networks are adjusted based on the conditions for optimal material distribution during training. Trained neural networks are showing excellent results to learn these dependencies \cite{wu2021deepcad,du2021rapid}. Thus, the NN can be trained to find new design variations which only consider functionality requirements. For this generative design processes, generative neural networks like Variational Autoencoder (VAE) \cite{Kingma.2013} and  Generative Adversarial Networks (GAN) \cite{Goodfellow.2014} are often successfully used. For generative neural networks based approaches functionality design requirements are assigned as conditions for a specific geometry. Newer approaches use Conditional Variational Autoencoder (CVAE) \cite{Sohn_2015} to generate objects under specific conditions. Training such a model with an increasing number of conditions is a major challenge. For this purpose, a low dimensional representation like a latent space is mostly used to represent multi functionality dependencies. It gives the opportunity to compare designs due to their similarities \cite{Shu.2020}. In this way, the chance is given to balance multi-criteria conditions for a higher-performance design. However, these latent spaces are difficult to interpret and analyze with a growing number of conditions. This is one reason why design problems solved so far with generative neural networks are mostly limited to a two-dimensional design space and a low number of considered conditions \cite{oh2019deep,chen2022inverse}. Further, existing approaches show lack of adaptation in the NN structure with the complexity of multiple conditions to learn the relationship between conditions and geometry. In particular, when conditions are strongly dependent, in the same geometry. Moreover, there are hardly any examples as a data basis for such 3D CAD generative neural networks use cases with major conditions. From this we derive our main research questions for this work: 
\begin{enumerate}
\item Can a generative neural networks be trained on 3D synthetic data with multi physics based conditions to generate a 3D design for a real design task?
\item How must the structure of a generative neural networks be extended to associate multi-criteria constraints during the training process with geometry features on a design?
\item How to train a generative neural network to predict more powerful designs and find an optimum under multiphysics conditions?
\end{enumerate}
To answer our research questions, we have a multi-criteria 3D design use case regarding an engine mount for an unmanned air vehicle. We generate 10,000 three dimensional generative designs, which is based on our earlier work \cite{Petroll.2021}. The designs are generated with respect to the functionalities and geometries features from our use case. With a total of 30,000 executed simulations we evaluate and label the generated generative designs in terms of its physics-based functionalities. The physics-based functionalities are mechanics, thermodynamics as well as aerodynamics. In addition, we introduce conditions, which enable an assessment of manufacturability with additive manufacturing. After, we train a CVAE with our labeled generative designs \cite{Sohn_2015}. A major challenge on our regression problem are the continuous values of our conditions. Moreover, the physical quantities values are numerically very different in scale. The introduction of more than one condition caused a large divergence in the latent space. It is an ambiguous learning behavior to concrete geometry features. Therefore, we semantically partition our conditions and extend the input NN structure of the CVAE. The trained model gives us the probability for a material prediction for all material areas in the design space. Finally, we use our trained model to generate an optimized design which fulfills best our multi-criteria functionalities. To the best of our knowledge, this is the first work that addresses the problem of a 3D design problem with regression multi-criteria conditions with a generative neural networks. Our contribution in this paper is threefold: 
\begin{itemize}
    \item An extended conditional variational autoencoder approach to open up a three-dimensional solution space with a geometrically parameter-free description of a component under multiple physics based conditions.
    \item An approach for a higher performance design generation, from a multi-condition learned relationship between latent representations and the generated designs.
    \item An evaluation of our presented approach on a 3D use case, with an interpretation of the latent space of a successfully trained generative neural networks for an optimal component design.
\end{itemize}

\section{Related Work}
\label{sec:SOTA}
\subsection*{Deep Learning for 3D Data}
In the field of computer vision, there has been a significant development of different deep networks for a variety of different tasks in recent years. For this reason, a variety of methods based on VAEs \cite{brock2016generative,kim2021setvae}, GANs \cite{gao2022get3d,wu2016learning}, diffusion models \cite{ye2022first,zeng2022lion,zheng2022neural} as well as normalising flows \cite{klokov2020discrete,liu2022let} have been explored to generate 3D objects as mesh, point cloud or voxel representations. So in the field of 3D object recognition to implement a joint embedding of 3D shapes and synthesised images approaches are shown in \cite{li2015joint,su2015multi}. Another approach is presented in \cite{sharma2016vconv} where the researchers used voxel-based models with an autoencoder to represent 3D objects. A more effective approach is used in \cite{qi2017pointnet} where a point-like representation is used to explore 3D objects. Other approaches like in \cite{yan2016perspective} use 2D images together with a 3D to 2D projection layer to generate 3D objects. Besides the classical use of the presented approaches for classification tasks \cite{sharma2016vconv,qi2017pointnet}, the approaches can also be used for completing full shapes \cite{chen2019unpaired,tchapmi2019topnet} or for single-view reconstruction \cite{mandikal20183d}. Furthermore, \cite{chen2019text2shape,fu2022shapecrafter} are exploring text-based 3D object generating approaches.

\subsection*{Conditional Variational Autoencoder}
\label{CVAE}
Based on the concept of a VAE \cite{Kingma.2013} a Conditional Variational Autoencoder \cite{Sohn_2015} (CVAE) is considered good to represent the high-dimensional joint distributions of features \cite{Sohn_2015,kim2021conditional,yonekura2021data}.\\
The main target of VAEs is the estimation of the relation between the input $x_i$ and the corresponding latent representation $z_i$. In variational inference, the posterior $p(z|x)$ is approximated by a parameterized distribution $q_{\theta}(z|x)$ called the variational distribution. The lower bound for $p(x)$ can be written as follows:
\begin{equation}
\label{eq:VAE}
L_{\theta,\phi,x}=E_{q_{\theta}(z|x)}\ [log\ p_{\phi}(x|z)] -KL(q_{\theta}(z|x) || p_{\phi}(z))
\end{equation}
The two fundamental parts of the VAE are the Encoder $E=q_{\theta}(x|z)$ with parameters $\theta$ and the Decoder $D=p_{\phi_{D}}(x|z)$ with parameters $\phi$. They represent functions which map the input $x_{i}$ to a latent space $z_{i}$ and vice versa. The reconstruction from $x_{i}$ is $\hat{x}_{i}$. Here, the represented optimization is an minimization of the reconstruction loss under consideration of the KL divergence as an regularizer. $E$ has two outputs $\mu_{i}$ and $\sigma_{i}$ that correspond to the mean and the standard deviation of the Gaussian latent variable $z_{i}$. For this, the reparameterization trick \cite{Kingma.2013} is normally used with  $\mu_{i}+\sigma_{i}*\epsilon$ under consideration of $\epsilon_{i}\sim N(0,1)$ to calculate $z_{i}$. It helps the network to shift not to much from the true distribution.\\
In contrast to the VAE, a CVAE approach based on the maximisation from the variational lower bound of the conditional likelihood $p(x|c)$ which supports to generate designs under multiple specified conditions $\textbf{c}=\{c_{1}\dots c_{n}\}$ where $n$ is the number of conditions.
\begin{equation}
\label{eq:CVAE}
\begin{split}
L_{\theta,\phi,x,\textbf{c}}
&=E_{q_{\theta}(z|x,\textbf{c})}\ [log\ p_{\phi}(x|z,\textbf{c})]\\
&\quad -KL(q_{\theta}(z|x,\textbf{c}) || p_{\phi}(z|\textbf{c}))
\end{split}
\end{equation}
The trained CVAE is usable to reconstruct an input $x_{i}$ under a set of conditions $\textbf{c}$ to match the target outputs $\hat{x}_{i}$. In contrast to the VAE, the main parts of the CVAE $E$ and $D$ are conditioned by $\textbf{c}$. It follows that $E=q_{\theta}(z|x,\textbf{c})$ with parameters $\theta$ and $D=p_{\phi_{D}}(x|z,\textbf{c})$ with parameters $\phi$ which represents functions are used to map the input $x_{i}$ under consideration of $\textbf{c}$ to a latent space $z_{i}$ and vice versa. In this context, a core problem is when working with multiple conditions in a CVAE is how to bring them into the network. Also the weighting or balancing problem of the reconstruction error and the Kullback-Leibler divergence shows this. It has been object of several investigations \cite{asperti2020balancing}.
\begin{figure}[t]
 \centering
 \begin{center}\includegraphics[width=12cm,height=18cm,keepaspectratio]{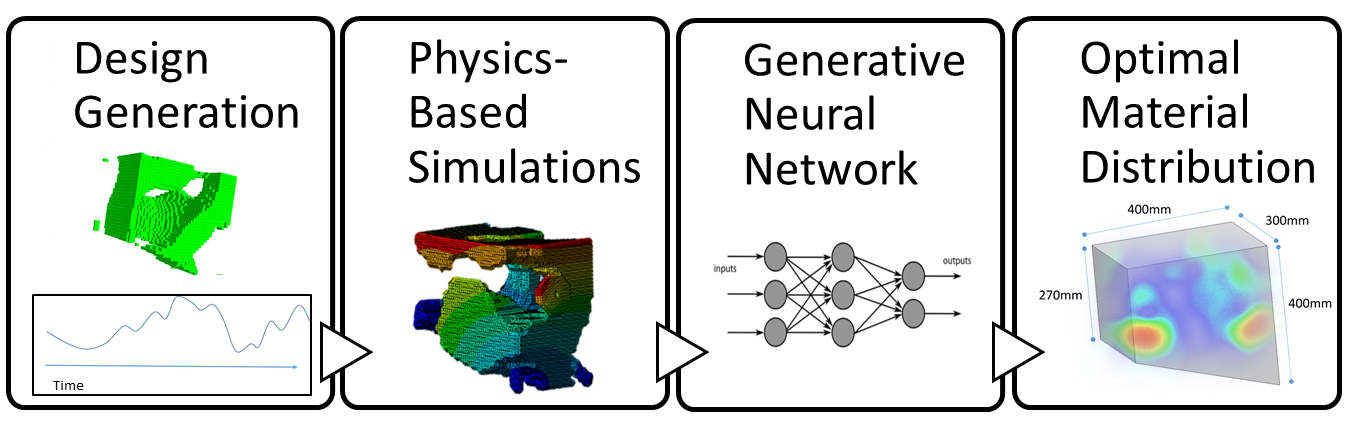}\end{center}
 \caption{Multi-criteria generative neural networks design approach. Four steps of using a Deep Input CVAE for functionalities based design generation.}
 \label{fig:4StepOverview.png}
\end{figure}

\subsection*{Deep Learning for Engineering Tasks}
For iterative design generation, in \cite{Shu.2020} a GAN based approach is shown for direct 3D modeling of an aircraft. The work followed the idea of a physics-based generated dataset. Thereby, the aerodynamics are considered primarily and the shape as the single condition. The goal is to minimize the aerodynamic drag. Furthermore, in \cite{Nobari.2021} an approach is developed for generating 3D models with more constraints. The researchers add a range loss, so design constraints are additionally taken into account based on parameter specifications using the example of 3D aircraft models. A slightly different approach is presented in \cite{Zhang.2019} for the optimization of 3D models. After successfully training of a variational autoencoder, a genetic algorithm is used to optimize the latent space design embeddings. Further,  an approach to consider continuous conditions in the generation process with Conditional GANs is shown in \cite{Nobariv.2021}. They use a singular vicinal loss in combination with a loss function based on determinant point processes. In doing so, the researchers add a new self-amplifying Lambert Log Exponential Transition Score, which is used for improved conditioning. They successfully demonstrate the approach on an 2D airfoil generation task with diverse results. Similarly, a Free-Form Deformation Generative Adversarial Networks which provides efficient parameterization for 3D shapes is presented in \cite{Chen.2021b}. Hereby, they achieve high representation compactness and capacity. A VAE to select an optimal material strength for their 2D optimization approach to retrieve a result from a latent space is shown in \cite{qian2020accelerating}. They take a structure optimization and determines the optimal material from the latent space of their trained VAE.\\
The work presented shows the difficulty of available data for design problems. Data for more complex solutions for multiple conditions isn't published. 3D data and corresponding physics-based labels are missing.  GAN approaches are available in detail mostly with one considered condition. Multiphysics problems are missing in the context of direct 3D design creation completely or don't deal with real physics-based designs \cite{uugur2019multi}. Further, it is recognisable that generative neural networks are often used for classification problems, which have not been further discussed here. In summary, an approach which allows to incorporate three-dimensional multi-criteria designs with regression conditions into a generative neural network is missing. Therefore, no extended NN approaches which have a change in their architecture in favor of multi-criteria conditions do exist.

\section{Method}
\label{sec:method}
In this section we propose our method to effectively bring continuous multi-criteria conditions into a new design of an UAV. In doing so, we solve a multi-physics design problem with a CVAE. The use case gives concrete functionalities and geometry features. Finally, a multi-physics and functionally generative optimal design is presented. Optimal with regard to the physical conditions.
To achieve this target, we developed a four-step approach to use a generative neural networks for a new design of a component (Figure: \ref{fig:4StepOverview.png}).\\ 

First, we generate 10,000 designs. We do it with a pseudo random noise based on \cite{bae2018perlin} and our earlier work \cite{Petroll.2021}. The special random function should ensure that the design space is covered completely and evenly \cite{Petroll.2021}. For the neural network each voxel should occur equally often in training. This is done to teach the neural network a parameter-free geometric generative description of design variants. In this way a geometric solution appears as unrestricted as possible in the design space (generative design). Second, we use a physics-based simulation to evaluate our designs with respect to the functionalities. Each generated generative design is labeled with its physical performance data. Third, we train a  generative neural networks with an extended architecture and our generated designs as well as simulation based labels as input. It learns where material in the design space is important or unimportant for the physics-based functionalities. In a last step the trained NN is used to generate an optimal design. At this point the lower dimensional latent representation is used for a prediction of a new design under regression multi-criteria conditions. The main differences of our approach to existing approaches is as follows. We show a generative neural networks based approach in which not only single criterion requirements for a design problem are solved. The design problem is three-dimensional, multi-physical and considers geometric requirements and interfaces. A concrete use case and additive manufacturing are also addressed. The extension of a CVAE is developed, demonstrated and improved for multi criteria conditions. 
\subsection{Training Data Generation}
To create the 10,000 generative designs $\textbf{X}=\{x_{1}\dots x_{i}\},\ i\in\{1\dots10,000\}$ we use a noise based generation method. We define our design space $\mathrm{A}$ with 50,400 voxels (Figure: \ref{eq:Design_space})
\begin{equation}
\label{eq:Design_space}
    \mathrm{A}=\{(\mathrm{a}_{j,k,l}) | \forall \ j=1,2,\dots30,\ k=1\dots40,\ l=1\dots42\}
\end{equation}
where one voxel $\mathrm{a}_{j,k,l}$ per $\mathrm{cm}^3$ is used. This comparatively rough representation is chosen due to the expected long computation and power calculation time. Next a three-dimensional Perlin Noise ($\mathrm{noise}$) is used to generate a basic material distribution $\mathrm{Md}_{\mathrm{AM},i}$ in the design space $\mathrm{A}$
\begin{equation}
\label{eq:Perlinnoise}
   \mathrm{Md}_{\mathrm{AM},i}=\sum_{n=0}^{M-1} \hat{u}*\mathrm{noise}(\nu_{n}*\mathrm{a}_{j,k,l})
\end{equation}
with amplitude modulation $\mathrm{AM}$, frequency $\nu$ and amplitudes $\hat{u}$. Here, $\hat{u}_{n+1}=\hat{u}_{n}*\phi_{\mathrm{noise}}$ is guilty where a combination of the frequency and amplitude modulation with different frequencies is used. At this point, $\phi_{\mathrm{noise}}$ is a special constant which links the amplitude with the amplitude of the previous step. This creates uniform coverage of the design space. At the correct scale it produces organic-looking designs due to the basis of locally contiguous duration's.\\
After, where the engine mount needs interfaces to the engine and to aircraft structure, material is used per design (Figure: \ref{eq:Design_space}). Through repetitive areas, the NN learns where in any case must be material for addon parts. Algorithms are used to ensure that the designs can be use for a physic-based simulation \cite{petroll2021synthetic}. So the design consists of only one body and can be flowed through by air\cite{petroll2021synthetic}. As a final step, the designs are transformed into a surface description for simulations. The described steps from Eq. \ref{eq:Perlinnoise} are repeated until a quantity of 10,000 generated designs $\textbf{X}$ is achieved.

\subsection{Physics Based Simulations}
\begin{figure}[h]
 \centering
 \begin{center}\includegraphics[width=12cm,height=18cm,keepaspectratio]{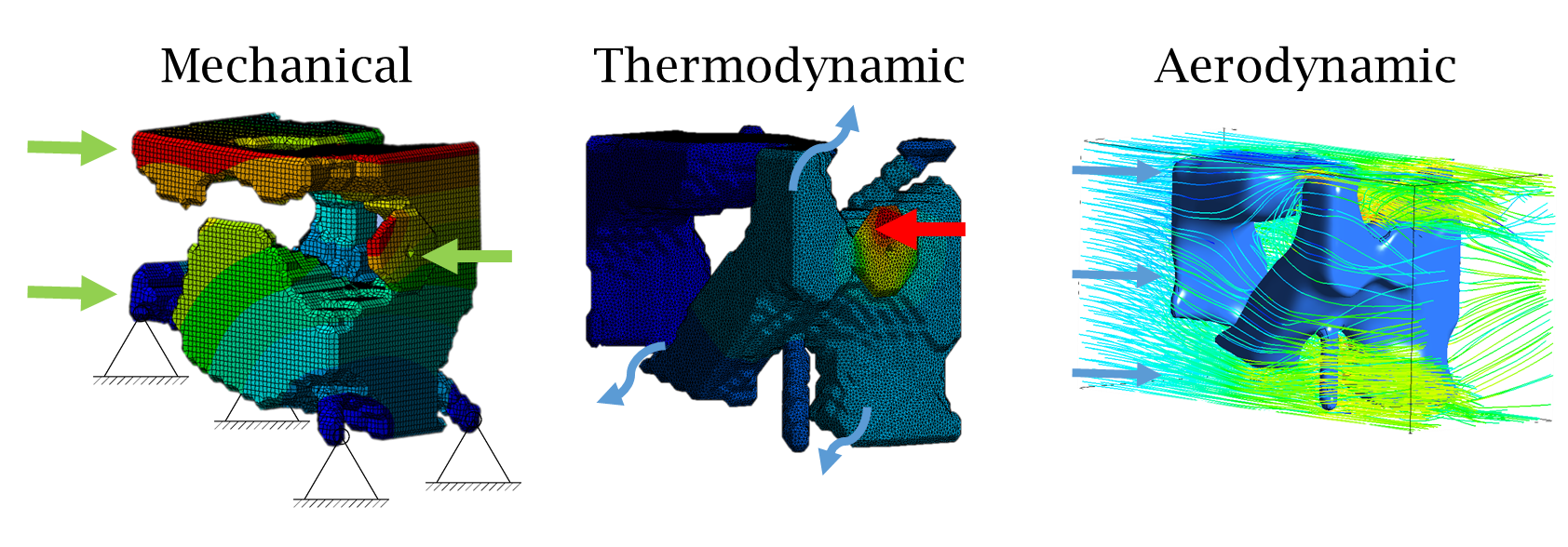}\end{center}
 \caption{ANSYS simulations for mechanics, thermodynamics and aerodynamics with generated designs and specific load cases. The arrows represent the direction of it.}
 \label{fig:Simulation.png}
\end{figure}
The simulation based label generation is done with automated simulations in ANSYS \cite{madenci2015finite} FEM and CFD (Figure: \ref{fig:Simulation.png}). For this purpose, one mechanical, one thermal and one aerodynamic simulation for each generated design $x_i$ is performed. 

We take these as the basis for our considered physic-based parameters, which are most expressive for our use case. So our conditions where $n=9$ are the following: \\
For the mechanics, we evaluate the mean residual stress $c_1$ and mean total deformation $c_2$ for all voxels. For thermodynamic, mean temperature $c_3$ and heat density $c_4$. In aerodynamics we consider the mean outlet pressure $c_5$ and the resistance to air $c_6$ in the direction of flow. For the previously mentioned conditions, we don't use maxima values cause of bad training tests. Instead, we use mean values, so that the value distribution corresponds more closely to a Gaussian distribution. The assessment with regard to additive manufacturing is based on the heat distribution in the printing process $c_7$ and build-up angle of number to surfaces $c_8$. A lightweight design criterion $c_9$ is introduced as a classic optimization factor. For this consideration we use the number of voxels per design. Finally we have a set 10,000 value pairs with one value for each condition per generated design $\textbf{C}_x=\{\textbf{c}_{x_1},\dots ,\textbf{c}_{x_{10000}}\}$. So, all values are normalized per condition for faster and better training. Values in the set of conditions $c_{n}=\{c_{n,1},\dots, c_{n,10000}\}$ which are not in range of $\displaystyle \pm\ 2 \sigma$ are dropped out. This leads to a more uniform training process, which does not focus on maximums. 

\subsection{Extended CVAE for Multi-Physics Based Design Generation}
\label{CVAEdddd}
\begin{figure*}[t]
\centering
\includegraphics[width=1\textwidth]{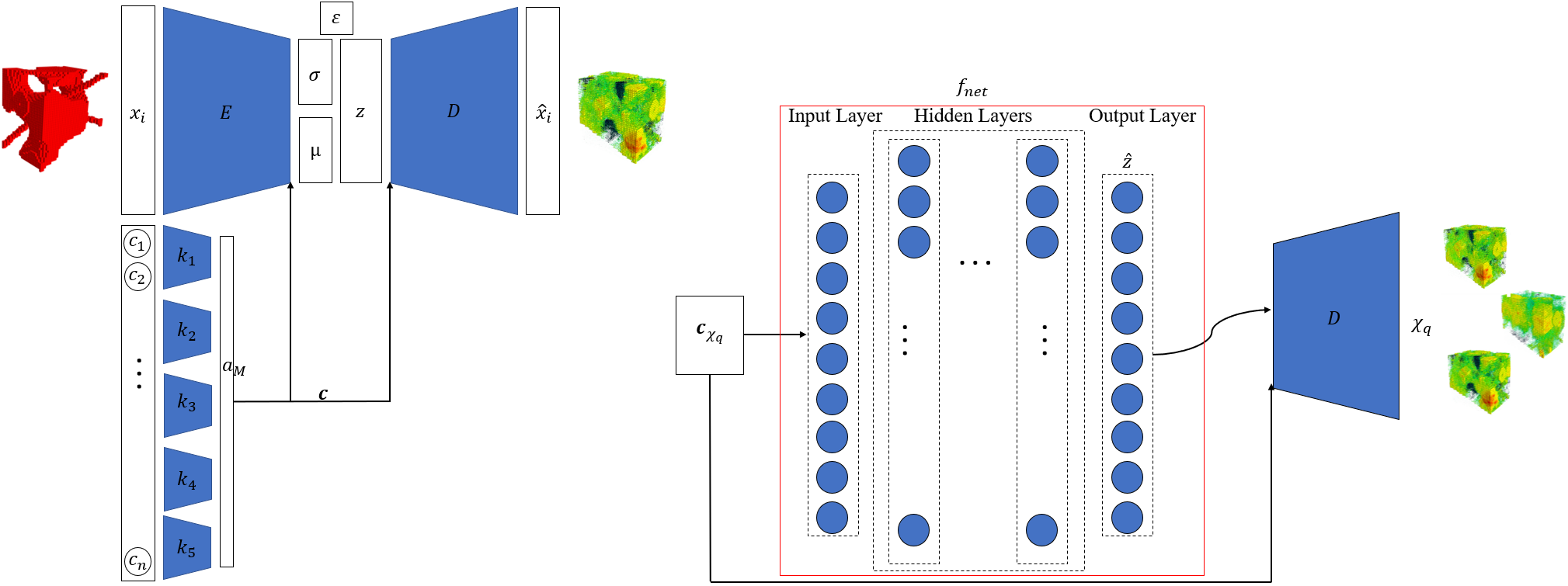} % Reduce the figure size so that it is slightly narrower than the column.
\caption{On the left our developed Deep Input CVAE approach for reconstruction of 3D object with multiple conditions $c_n$ and $k_{1,\dots,5}$ is illustrated. On the right our approach for design generation with a FNN as well as the trained decoder $D$ is shown. The data flow as well as the neural network structure is shown to generate new designs $\chi_q$ with self selected values $c_\{\chi_q\}$. In red, $f_{\mathrm{net}}$ is highlighted.}
\label{fig:NewApproach.png}
\end{figure*}
We use as a generative neural network a CVAE like in \ref{CVAE} explained with the generated designs $\textbf{X}$ and the values of $\textbf{C}_x$ as input. With data augmentation like in \cite{kar20223d} we support our training. Further we use seven dense layers to reduce the size as follows ($50400,1024,512,..., 16$) as well as in reverse order in $D$. After each layer we use a Rectified Linear Unit (ReLU) activation function The latent size $z$ is defined as 32. Further we apply an Adam optimizer \cite{kingma2014adam}. For the generation of $q(z)$ as close as possible to the standard normal distribution we are using a two-part loss function with the reconstructions loss $E[\cdot]$ as well as a KL-divergence loss $KL[\cdot]$ like in Eq.\ref{eq:CVAE}.\\
In comparison to the state of art, we divide $\textbf{c}$ given the physic discipline of each $c_{n}$ into five categories $\textbf{K}=\{k_{1}\dots k_{b}\}, b \in \{1.\dots,5\}$. In doing so, we have used the following allocations: $k_{1}=\{c_{1},c_{2}\}$, $k_{2}=\{c_{3},c_{4}\}$, $k_{3}=\{c_{5},c_{6}\}$, $k_{4}=\{c_{7},c_{8}\}$ and $k_{5}=\{c_{9}\}$. This extension supports the combination of values which differ significantly in their dimensions. In this context, we use complementary simple feedforward neural network (FNN) structure extension for each category in the input of our Deep Input CVAE (D-CVAE). The five extensions are added in one layer $a_{\textbf{M}}$ after seven hidden layers size ($4,8,...,256$) per extension $k_{b}$ and concatenated with last layer of $E$ and the first layer of $D$. This semantically separated and more complex representation of our input improves the representation of the complex data strongly. After training the D-CVAE, trained $D$ represents a parametric model where $z_i$ and $\textbf{c}_i$ are input parameters to generate new designs. So, an opportunity is given with the trained D-CVAE to generate a new design $\chi_{q}$ with the desired performance maximization across all conditions. The D-CVAE architecture previously described is shown in Figure \ref{fig:NewApproach.png}.

\subsection{Design Optimization}
An approach for an optimal design generation $\chi_{\mathrm{opt}}$ follows on. First to generate an a new design $\chi_{q}$ with self selected values for each condition in $\textbf{c}$, the relationship between the latent representation per design $z_{{x_i}}$ and $\textbf{c}_{{x_1}}$ is trained. For this purpose a FNN ($f_{\mathrm{net}}$) is used to learn this relationship to predict $\hat{z}_{{x_i}}$ as a new latent representation:
\begin{equation}
\label{eq:new_Z}
\hat{z}_{i}=f_{\mathrm{net}}(\textbf{c}_i)
\end{equation}
In- and output variables to train the FNN $f_{\mathrm{net}}(\textbf{c})$ with eight hidden layers and a ReLU activation function to predict a latent representation per generated design $\hat{z}_{{x_i}}$ are $\textbf{c}_{{x_1}}$ as well as $z_{{x_i}}$. The trained $f_{\mathrm{net}}$ allows with trained $D$ and self selected values $\textbf{c}_{\chi_{q}}$ for $\textbf{c}$ to predict a permissible quantity $q$ (e.g. $q=100$) of new individual models $\mathrm{Q}=\{\chi_{1},\dots,\chi_{q}\}$. Therefore the following applies under consideration of $\textbf{c}_{\chi_{q}}$
\begin{equation}
\label{eq:new_modell}
\chi_{i}=D(\hat{z}_{i},\textbf{c}_{\chi_{q}}).
\end{equation}
The explained approach to generate $\chi_{q}$ is shown on the right in Figure: \ref{fig:NewApproach.png}.\\
Once $f_{\mathrm{net}}$ is successfully trained, $\hat{z}_{i}$ can be used in a three step way to determine an optimal design $\chi_{\mathrm{opt}}$ in performance across all conditions.\\
First, the values of every physical property $\textbf{c}_{{n,x}_{i}}$ are ordered per condition from the user's point of view from the lowest to the maximum performance. This is given by the minimum and maximum performance by the value range per condition. This creates new value pairs $\textbf{C}_{\chi}=\{ \textbf{c}_{\chi_{q}},\dots, \textbf{c}_{\chi_{q}}\}$. These value pairs are are not previously present in $\textbf{C}_x$. The new value pairs are the basis for more powerful designs. Thereby, $\chi_{1}$ has the lowest performance for all $\mathrm{c}_{n,\chi_{q}}$ while $\chi_{q}$ has the highest performance per $\mathrm{c}_{n,\chi_{q}}$ so

\begin{equation}
\label{eq:100Designs}
\begin{split}
\chi_{1}=D(\hat{z}_{1},\textbf{c}_{\chi_{1}}) <  \chi_{2}=D(\hat{z}_{2},\textbf{c}_{\chi_{2}}) <  \dots \\< \chi_{q}=D(\hat{z}_{q},\textbf{c}_{\chi_{q}})
\end{split}
\end{equation}
is guilty.

Next, to see if we can push our self-selected values even further to a higher performance design, we look at the variety which our D-CVAE can provide. For this we use the material change rate $\Delta M$ of each design point per design $\chi_{q}$ to the next $\chi_{q+1}$ calculated by
\begin{equation}
\label{eq:MaterialChangeRate}
\Delta M = \sum_{1}^{q-1}\frac {\sum_{1}^{j}\sum_{1}^{k}\sum_{1}^{l}(A_{j*k*l})_{\chi_{q+1}} -(A_{j*k*l})_{\chi_{q}})}{j*k*l}.
\end{equation}

We use Eq.\ref{eq:MaterialChangeRate} to define the range where the new values for our defined conditions can be set. We assume that a new value per condition can only be set in the area where trained $D$ has enough diversity in the design. This defines the limit for our trained model to retrieve a design with maximum performance from the latent space. Finally, we analyze the point where the material change rate is maximum while considering maximum performance. This results in the best possible design Eq. (\ref{eq:NewGradient}) with the presented optimization approach.
\begin{equation}
\label{eq:NewGradient}
    \chi_{\mathrm{opt}}: max\ f(\Delta M) \ \text{for} \: min\ D(\hat{z}_{i},\textbf{c}_{\chi_{q}}) 
\end{equation}
Here, $\chi_{\mathrm{opt}}$ defines the optimal material distribution for a higher performance design. For a simulation based validation $\chi_{opt}$ has to be transformed manually via \cite{lorensen1998marching} to a CAD file.

\section{Experiments}
\label{sec:experiments}

In this section, we report the details of our experiments and the qualitative and quantitative validation. We compare our approach with a 3D Convolutional Neural Network in conjunction with a CVAE (CNN-CVAE) presented by \cite{na2018toxic} and a fully connected layer (FC-CVAE) presented in \cite{canchumuni2019towards}. In addition, we show the results in terms of an optimal design generation.

\subsection{Use Case}
\begin{figure}[h]
 \centering
 \begin{center}\includegraphics[width=12cm,height=18cm,keepaspectratio]{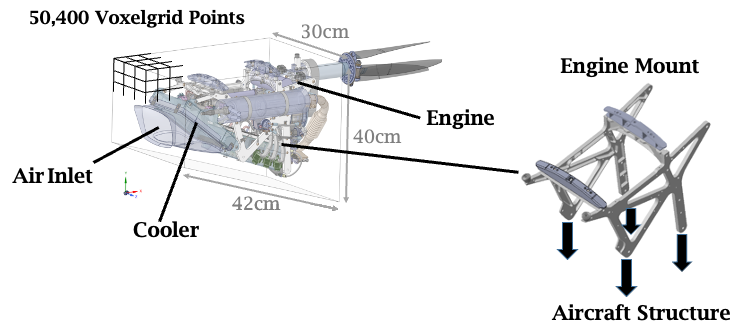}\end{center}
 \caption{On the right is the complete UAV drive unit and on the left the engine Mount. As a 3D data representation method a voxel-based geometric description is used.}
 \label{fig:Engine.png}
\end{figure}
Our use case is the design of an engine mount for an unmanned air vehicle (UAV) displayed in Figure \ref{fig:Engine.png}). The idea is to reduce the number of components as far as possible to one central design with add-on parts (e.g. electronic, engine). For this purpose, the new possibilities of additive manufacturing are considered. \\
To achieve the target, we analyze the engine in terms of its main functionalities. In this case, the Wankel engine is attached to an engine mount that transmits the thrust to the aircraft structure. For operation, there is a radiator at the beginning of the engine, which cools the engine through coolant pipes located on the engine mount. In the particular case of the launch phase on a catapult, much heat is transported from the engine in the engine mount.\\

The main functionality of the engine mount can be described as the static stability to hold and sufficient heat dissipation to cool the engine. To ensure these functionalities, air must flow freely through the engine mount.
Our goal is derived from this to design with a CVAE a holder which can withstand the mechanical and thermal load case, and has a favorable aerodynamic design. In addition, conditions which make metal additive manufacturing feasible must be considered. 
\subsection{Training Settings}
\begin{figure}[h]   
 \centering
 \begin{center}\includegraphics[width=12cm,height=18cm,keepaspectratio]{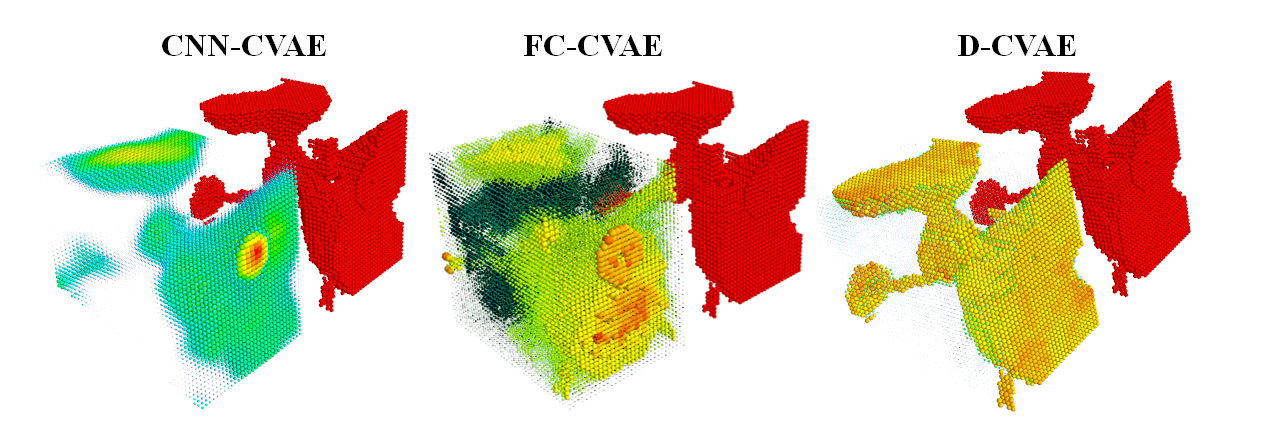}\end{center}
 \caption{Reconstruction of the models depending on the condition. On the right side in red the original and on the left in colors representing the probability of material.}
 \label{fig:Designs.png}
\end{figure}

The Training is done on a Xeon 4108 with 64GB RAM and 1 GPU NVIDIA P5000. Training results for the mentioned types generative neural networks are shown in Figure (\ref{fig:Designs.png}). It can be seen for multiple conditions the reconstruction result for our designs becomes more and more fuzzy. First, when using a CNN-CVAE compared to the FC-CVAE the core body of the design is presented well.
\begin{table}[h]
\begin{center}
\begin{tabular}{|l|r|r|r|}
\hline
Model & KL-Loss &Reconstruction loss &Total \\
\hline
CNN-CVAE & 90 & 6,050 & 6,140\\
FC-CVAE & 2,500   & 30,000 & 32,500\\
\textbf{D-CVAE}  &\textbf{80} &\textbf{520}  & \textbf{600} \\
\hline
\end{tabular}
\caption{Absolute final values of the loss functions as well as the total loss after training per considered generative neural network-approach.}
\label{tab:loss}
\end{center}
\end{table}
It is noticeable that subtleties are not present in the generated designs. Also, the training time is 8h for 200 epochs. Therefore, hyperparameter tuning is very time consuming. Compared to a training time of 40 minutes, the FC-CVAE is much faster, but it shows a very noisy design. Interesting is the observation of areas where material is very unlikely which is displayed numeric negatively (dark blue areas in Figure \ref{fig:Designs.png}).\\
In the following Table \ref{tab:loss} our final loss values are presented. The reconstruction loss can be understood as the number of misrepresented voxels. The KL Loss is a measure for the quality of the conditions learned. 
\begin{figure}[h]
\centering
\begin{center}\includegraphics[width=12cm,height=18cm,keepaspectratio]{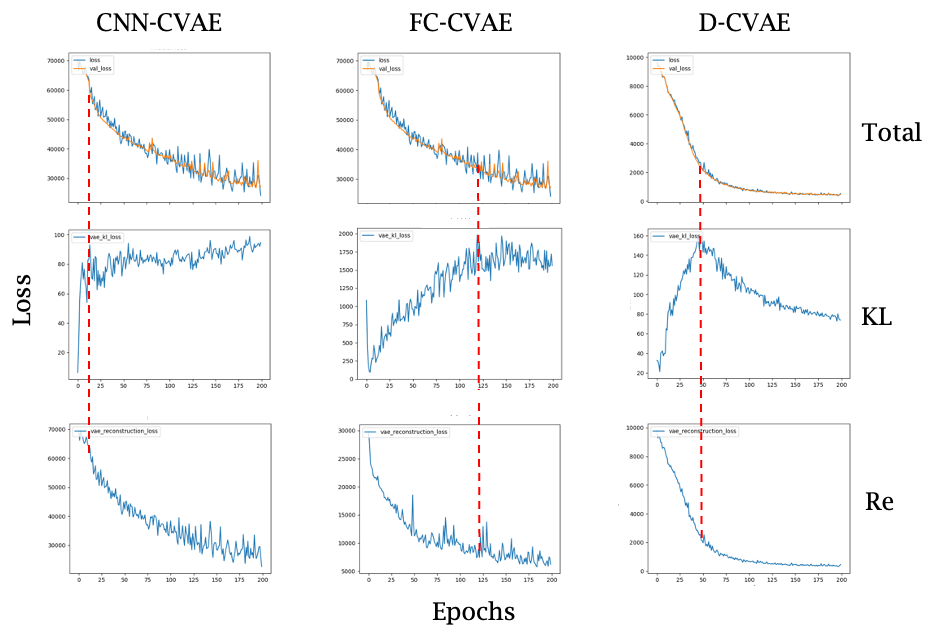}\end{center}
\caption{The loss functions of the three trained models CNN-CVAE, FC-CVAE and our defined model D-CVAE are shown here. Top total loss, mid KL loss and bottom reconstruction loss. The red line points to the maximum in the KL loss and the adjustment of the reconstruction loss. It is recognizable that our developed D-CVAE has the best training curves.}
\label{fig:Lernkurven.png}
\end{figure}
Our D-CVAE shows a natural balancing of our label to learn the latent space. This is shown in Figure \ref{fig:Lernkurven.png}. The total loss with of the D-CVAE improves significantly compared to the CNN-CVAE and FC-CVAE and the two loss components. KL Loss and Reconstruction Loss, converge (redline) by themselves in such a way that the conditions have a sufficient influence.

\begin{table}[h]
\begin{center}
\begin{tabular}{ |l|r|r| }
\hline
Model & Abs. Error Design Space \\
\hline
CNN-CVAE & 1,762  \\
FC-CVAE & 2,625  \\
\textbf{D-CVAE} & \textbf{60}  \\
\hline
\end{tabular}
\caption{Abs. error in design space of design predictions $\hat{x}_i$.}
\label{tab:absolut}
\end{center}
\end{table}
On the basis of the representation of the learned design (Figure \ref{fig:Designs.png} and Table \ref{tab:absolut}) our D-CVAE approach generates qualitatively and quantitatively better results than the CNN-CVAE and FC-CVAE approaches. The D-CVAE shows the highest accuracy when it comes to mapping the contour. By adding up the probabilities of the predictions, the reliability of the predictions of a geometry from the condition can be determined by variance in Table \ref{tab:final}.

\subsection{Multi-Criteria Optimization of a 3D Design}

After we specify our own values per condition $\mathrm{c}_{n,\chi}$ to generate new models in the interest of design optimization. We create $\chi = 100$ values per condition $c_{n,\chi}$ from good to bad in the sense of our used case and the performance.

The following method is used to select the values: Thermals and mechanics should withstand the loads as much as possible and are demanded as constant conditions. Aerodynamics and manufacturability should improve over the 100 labels from 0-100. The ninth condition ($c_9$), which should ensure that less material is used, as a classic optimization requirement. $Min.$ and $Max.$ from the simulated conditions are used as upper and lower limits. The challenge here, is that condition combinations are now required which are not previously learned in the latent space. In total these are 100 new values pairs $uzk$ of unknown designs. These hundred conditions are used to retrieve the desired designs in the form of material distributions from the latent space with the decoder $D(\hat{z}_{q},\textbf{c}_{\chi_{q}})$.
\begin{figure}[h]
 \centering
 \begin{center}\includegraphics[width=12cm,height=18cm,keepaspectratio]{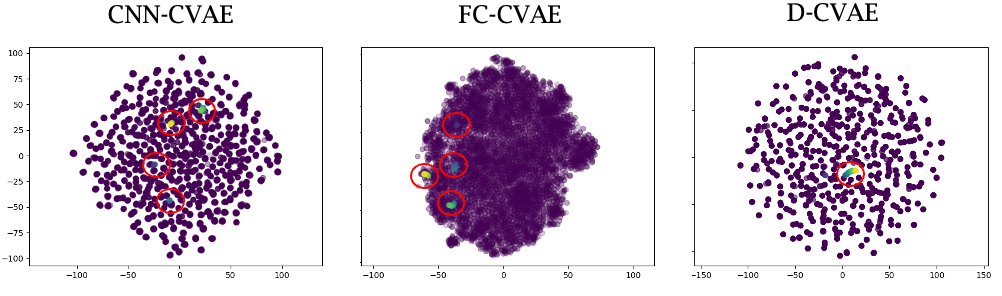}\end{center}
 \caption{Comparison of the latent space representation of the t-dispersed stochastic neighborhood implantation of the 100 user wanted performance values. We can see that the latent space of our D-CVAE has the best spread data representation.}
 \label{fig:Latentspace}
\end{figure}
At this point, we use the term material distribution instead of design proposal in the following, because the strongly competing nine conditions lead to the fact that no distinct design for arbitrary condition combinations can emerge clearly. Unfortunately, for our validation with simulations, this means that each material prediction with the new conditions has to be reconstructed manually. Thus, 4 examples each are chosen and simulated evenly split between 0-100 to look at the variance. From this, the variance $\sigma$ to the expected value $c_{\chi_{i}}$ is calculated from the given 100,\ $q \in Q$ conditions under consideration of $\sigma^2 = \sum_{1}^{q}(\mu-c_{\chi_{i}})^2$.

In addition to the actual values in Table \ref{tab:final}, there is an evaluation of how the value has developed from good to bad qualitatively. In comparison how it should develop according to the wanted performance per condition. Here, $\uparrow$ represents a qualitative improvement, $\downarrow$ on the other hand a degradation, $\rightarrow$ a preserve of the condition. The target performance defines how a better performing design should behave.
The D-CVAE shows the best results in terms of the qualitative consideration of the conditions. It can be seen that the D-CVAE tends to have lower variance in its simulated predictions than the other models (Table: \ref{tab:final}).\\
\begin{table*}[h]
\begin{center}
\begin{tabular}{|l|r|r|r|r|r|r|r|r|r|}
\hline
Modell & \multicolumn{1}{c|}{$c_1$} &\multicolumn{1}{c|}{$c_2$} &\multicolumn{1}{c|}{$c_3$} &\multicolumn{1}{c|}{$c_4$} &\multicolumn{1}{c|}{$c_5$ }&\multicolumn{1}{c|}{$c_6$} &\multicolumn{1}{c|}{$c_7$ }&\multicolumn{1}{c|}{$c_8$ }&\multicolumn{1}{c|}{$c_9$} \\
\hline
VAR CNN-CVAE & 0.210    &0.30    &261   &0.07 &1,202  &569 &9.07 &1,047 &0.07\\
VAR FC-CVAE & 0.150    &0.31   &169   &0.09 &2,015     &2,000  &14.05  &476  &0.10\\
\textbf{VAR D-CVAE} & 0.077    &0.27   &62   &0.03 &939    & 2,030  &7.22  &501  &0.13\\
\hline \centering
CNN-CVAE &\multicolumn{1}{c|}{$\rightarrow$} &\multicolumn{1}{c|}{$\downarrow$ }&\multicolumn{1}{c|}{$\rightarrow$}  &\multicolumn{1}{c|}{ $\rightarrow$} &\multicolumn{1}{c|}{$\nearrow$} &\multicolumn{1}{c|}{ $\uparrow$} &\multicolumn{1}{c|}{ $\nearrow$ }&\multicolumn{1}{c|}{ $\nearrow$ }&\multicolumn{1}{c|}{ $\uparrow$ }\\
FC-CVAE & \multicolumn{1}{c|}{$\rightarrow$} &\multicolumn{1}{c|}{$\rightarrow$} &\multicolumn{1}{c|}{$\searrow$}  &\multicolumn{1}{c|}{ $\searrow$} &\multicolumn{1}{c|}{$\rightarrow$} &\multicolumn{1}{c|}{$\nearrow$} &\multicolumn{1}{c|}{$\rightarrow$ } &\multicolumn{1}{c|}{$\uparrow$} &\multicolumn{1}{c|}{$\uparrow$}\\
\textbf{D-CVAE} &\multicolumn{1}{c|}{$\rightarrow$}  &\multicolumn{1}{c|}{$\rightarrow$} &\multicolumn{1}{c|}{$\rightarrow$} &\multicolumn{1}{c|}{$\rightarrow$} &\multicolumn{1}{c|}{$\nearrow$ }&\multicolumn{1}{c|}{$\nearrow$ } &\multicolumn{1}{c|}{$\uparrow$ } &\multicolumn{1}{c|}{$\uparrow$ } &\multicolumn{1}{c|}{$\uparrow$}\\
\hline
Target performance& \multicolumn{1}{c|}{$\rightarrow$ }&\multicolumn{1}{c|}{$\rightarrow$  } &\multicolumn{1}{c|}{$\rightarrow$ }  &\multicolumn{1}{c|}{$\rightarrow$} &\multicolumn{1}{c|}{$\uparrow$ }&\multicolumn{1}{c|}{$\uparrow$ } &\multicolumn{1}{c|}{$\uparrow$ } &\multicolumn{1}{c|}{$\uparrow$}  &\multicolumn{1}{c|}{$\uparrow$}\\
\hline
\end{tabular}
\caption{Variance (VAR) from predicted and simulated designs in comparison to the used condition. Together qualitatively presented with the desired performance that should be achieved}
\label{tab:final}
\end{center}
\end{table*}
The 100 desired labels can be seen in the latent space of the models in Figure (\ref{fig:Latentspace}). It is recognizable that a clearer range in the D-CVAE appears. The area in which the 100 conditions are retrieved is contiguous (red rings). The 100 points are marked from blue (poor performance) to yellow (good performance). Also, the distribution of the data shows more clearly distributed and separated points, which is indicative of a more diverse learned latent space. 
\begin{table}[h]
\begin{center}
\begin{tabular}{|l|c|c|c|}
\hline
Condition & Training& Our Opt.&Dev.[\%] \\
\hline
$c_1[MPa]$ & 0.062& 0.10&-48 \\
\hline
$c_2[mm]$ & 0.0032& 0.0072&-56 \\
\hline
$c_3[K]$ &216 & 291&+26 \\
\hline
$c_4[\frac{kW}{s}]$ &0.047 &0.095 &-51 \\
\hline
$c_5[Pa]$ & 198&26 & +716 \\
\hline
$c_6[N]$ &116 &52 &+223 \\
\hline
$c_7[\frac{mm^2}{layer}]$ &7.175 &2.300 &+311 \\
\hline
$c_8[surfaces]$ &1,317 & 795 & +165 \\
\hline
$c_9[\varnothing{\ voxel}]$ &0.234 &0.160 &+146\\
\hline
\end{tabular}
\caption{Our optimum compared in percent to the model with the best performance in our training's data set. The results for the mechanical and thermal load case remain the same as intended and keep the conditions. The other conditions improve significantly.}
\label{table:distribution}
\end{center}
\end{table}
Finally, we want to find the best possible solution for our 3D multi-criteria design problem. The goal is to find a material distribution in the design space that maximizes performance considering the conditions. But there are natural limits to retrieving better and better design proposals from our model. To find them we look at the range in which our model still shows sufficient diversity material prediction with respect to the conditions. For this we use the material change gradient from one design point to the next for our 100 created conditions (Eq. \ref{eq:MaterialChangeRate}).
\begin{figure}
 \centering
 \begin{center}\includegraphics[width=12cm,height=18cm,keepaspectratio]{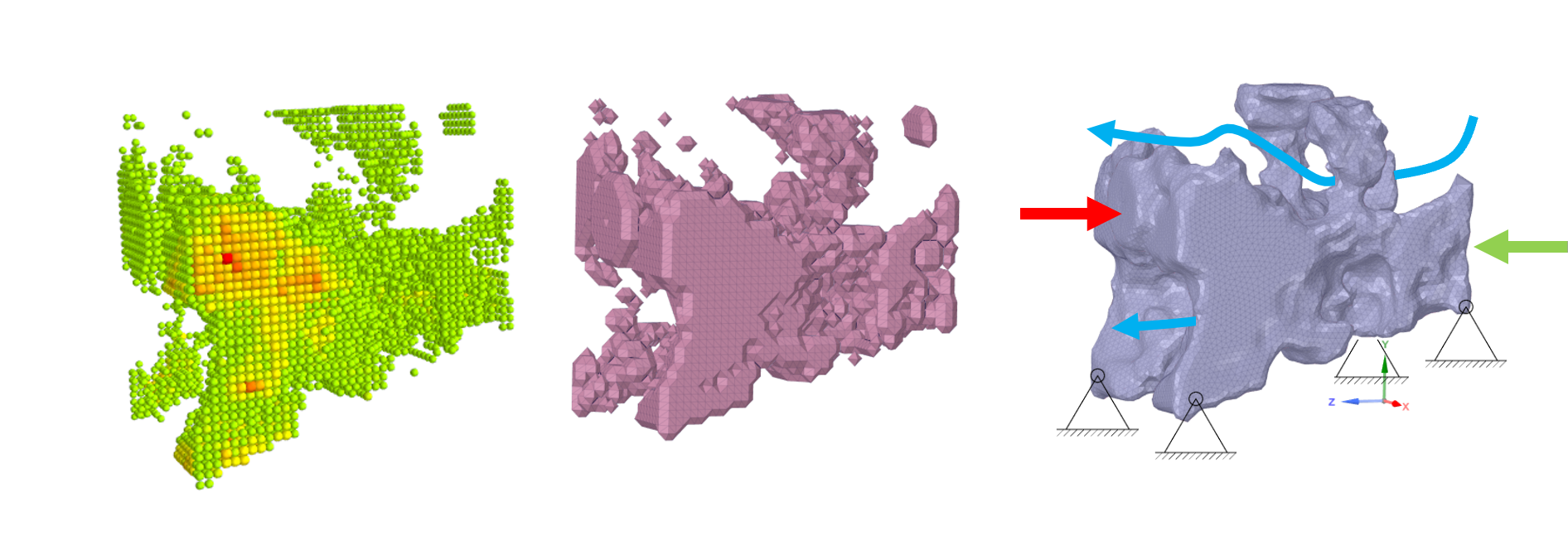}\end{center}
 \caption{Optimal material distribution and reconstructed CAD design. Arrows indicate the simulated load case for validation.}
 \label{fig:Optimalmaterial}
\end{figure}
The range in which significant material change can still be predicted for high performance is of interest. We searched in that manner for the best design with our trained model. We chose one recognizable maximum in the material change rate close to the maximum performance. The results are illustrated in Figure \ref{fig:Optimalmaterial}, it shows the product of a condition point. We simulate our optimum in all physical disciplines cf. Table \ref{table:distribution} and compare it to the best in our previously generated training data. From a qualitative point of view the results makes sense. There is a lot of material around the engine to remove the heat, there is a solid connection to the mounting points. In addition, the engine is directly surrounded by cooling air from two sides. The optimal model we generated is much better compared to the best model in the training data set.

\section{Conclusion}
\label{sec:conclusion}
The presented approach is one way to generate a 3D parameter-free geometry for a real multi-physics design problem with a D-CVAE. The main problem of determining a material distribution and linking geometry features to multiple regression conditions is met. However, the automated evaluation of the generated designs with D-CVAE is still a major obstacle for such a complex use case. It must be inferred repeatedly from to the material distribution to the design, similar to a classical topology optimization result. 
In this work one best design is shown in terms of our conditions as an optimum. It can be achieved without ground truth with the help of synthetic data. We have no comparison to an optimized component for all criteria with another method yet. Simpler data would not adequately address the complex challenge of multi-criteria design generation. Therefore, in further work, we concentrate on completely different conditional generative neural network approaches and new ways to clearly generate designs with multiple conditions. So, a faster and automated evaluation can be done with our data.
The data and code are available upon reasonable request.

\bibliographystyle{unsrt}  
\bibliography{references}  %%% Remove comment to use the external .bib file (using bibtex).

\begin{thebibliography}{10}

\bibitem{Jang_2022}
Seowoo Jang, Soyoung Yoo, and Namwoo Kang.
\newblock Generative design by reinforcement learning: Enhancing the diversity of topology optimization designs.
\newblock {\em Computer-Aided Design}, 146:103225, may 2022.

\bibitem{wu2021deepcad}
Rundi Wu, Chang Xiao, and Changxi Zheng.
\newblock Deepcad: A deep generative network for computer-aided design models.
\newblock In {\em Proceedings of the IEEE/CVF International Conference on Computer Vision}, pages 6772--6782, 2021.

\bibitem{du2021rapid}
Xiaosong Du, Ping He, and Joaquim~RRA Martins.
\newblock Rapid airfoil design optimization via neural networks-based parameterization and surrogate modeling.
\newblock {\em Aerospace Science and Technology}, 113:106701, 2021.

\bibitem{Kingma.2013}
Diederik~P Kingma and Max Welling.
\newblock Auto-encoding variational bayes.
\newblock {\em arXiv preprint arXiv:1312.6114}, 2013.

\bibitem{Goodfellow.2014}
Ian Goodfellow, Jean Pouget-Abadie, Mehdi Mirza, Bing Xu, David Warde-Farley, Sherjil Ozair, Aaron Courville, and Yoshua Bengio.
\newblock Generative adversarial nets.
\newblock {\em Advances in neural information processing systems}, 27, 2014.

\bibitem{Sohn_2015}
Kihyuk Sohn, Xinchen Yan, and Honglak Lee.
\newblock Learning structured output representation using deep conditional generative models.
\newblock In {\em Proceedings of the 28th International Conference on Neural Information Processing Systems - Volume 2}, NIPS'15, page 3483–3491, Cambridge, MA, USA, 2015. MIT Press.

\bibitem{Shu.2020}
Dule Shu, James Cunningham, Gary Stump, Simon~W. Miller, Michael~A. Yukish, Timothy~W. Simpson, and Conrad~S. Tucker.
\newblock 3d design using generative adversarial networks and physics-based validation.
\newblock {\em Journal of Mechanical Design}, 142(7), 2020.

\bibitem{oh2019deep}
Sangeun Oh, Yongsu Jung, Seongsin Kim, Ikjin Lee, and Namwoo Kang.
\newblock Deep generative design: Integration of topology optimization and generative models.
\newblock {\em Journal of Mechanical Design}, 141(11), 2019.

\bibitem{chen2022inverse}
Qiuyi Chen, Jun Wang, Phillip Pope, Mark Fuge, et~al.
\newblock Inverse design of two-dimensional airfoils using conditional generative models and surrogate log-likelihoods.
\newblock {\em Journal of Mechanical Design}, 144(2), 2022.

\bibitem{Petroll.2021}
Christoph Petroll, Martin Denk, Jens Holtmannsp{\"o}tter, Kristin Paetzold, Philipp H{\"o}fer, et~al.
\newblock Synthetic data generation for deep learning models.
\newblock In {\em DS 111: Proceedings of the 32nd Symposium Design for X (DFX2021)}, pages 1--10, 2021.

\bibitem{brock2016generative}
Andrew Brock, Theodore Lim, James~M Ritchie, and Nick Weston.
\newblock Generative and discriminative voxel modeling with convolutional neural networks.
\newblock {\em arXiv preprint arXiv:1608.04236}, 2016.

\bibitem{kim2021setvae}
Jinwoo Kim, Jaehoon Yoo, Juho Lee, and Seunghoon Hong.
\newblock Setvae: Learning hierarchical composition for generative modeling of set-structured data.
\newblock In {\em Proceedings of the IEEE/CVF Conference on Computer Vision and Pattern Recognition}, pages 15059--15068, 2021.

\bibitem{gao2022get3d}
Jun Gao, Tianchang Shen, Zian Wang, Wenzheng Chen, Kangxue Yin, Daiqing Li, Or~Litany, Zan Gojcic, and Sanja Fidler.
\newblock Get3d: A generative model of high quality 3d textured shapes learned from images.
\newblock {\em Advances In Neural Information Processing Systems}, 35:31841--31854, 2022.

\bibitem{wu2016learning}
Jiajun Wu, Chengkai Zhang, Tianfan Xue, Bill Freeman, and Josh Tenenbaum.
\newblock Learning a probabilistic latent space of object shapes via 3d generative-adversarial modeling.
\newblock {\em Advances in neural information processing systems}, 29, 2016.

\bibitem{ye2022first}
Mao Ye, Lemeng Wu, and Qiang Liu.
\newblock First hitting diffusion models.
\newblock {\em arXiv preprint arXiv:2209.01170}, 2022.

\bibitem{zeng2022lion}
Xiaohui Zeng, Arash Vahdat, Francis Williams, Zan Gojcic, Or~Litany, Sanja Fidler, and Karsten Kreis.
\newblock Lion: Latent point diffusion models for 3d shape generation.
\newblock {\em arXiv preprint arXiv:2210.06978}, 2022.

\bibitem{zheng2022neural}
Yan Zheng, Lemeng Wu, Xingchao Liu, Zhen Chen, Qiang Liu, and Qixing Huang.
\newblock Neural volumetric mesh generator.
\newblock {\em arXiv preprint arXiv:2210.03158}, 2022.

\bibitem{klokov2020discrete}
Roman Klokov, Edmond Boyer, and Jakob Verbeek.
\newblock Discrete point flow networks for efficient point cloud generation.
\newblock In {\em European Conference on Computer Vision}, pages 694--710. Springer, 2020.

\bibitem{liu2022let}
Xingchao Liu, Lemeng Wu, Mao Ye, and Qiang Liu.
\newblock Let us build bridges: Understanding and extending diffusion generative models.
\newblock {\em arXiv preprint arXiv:2208.14699}, 2022.

\bibitem{li2015joint}
Yangyan Li, Hao Su, Charles~Ruizhongtai Qi, Noa Fish, Daniel Cohen-Or, and Leonidas~J Guibas.
\newblock Joint embeddings of shapes and images via cnn image purification.
\newblock {\em ACM transactions on graphics (TOG)}, 34(6):1--12, 2015.

\bibitem{su2015multi}
Hang Su, Subhransu Maji, Evangelos Kalogerakis, and Erik Learned-Miller.
\newblock Multi-view convolutional neural networks for 3d shape recognition.
\newblock In {\em Proceedings of the IEEE international conference on computer vision}, pages 945--953, 2015.

\bibitem{sharma2016vconv}
Abhishek Sharma, Oliver Grau, and Mario Fritz.
\newblock Vconv-dae: Deep volumetric shape learning without object labels.
\newblock In {\em Computer Vision--ECCV 2016 Workshops: Amsterdam, The Netherlands, October 8-10 and 15-16, 2016, Proceedings, Part III 14}, pages 236--250. Springer, 2016.

\bibitem{qi2017pointnet}
Charles~R Qi, Hao Su, Kaichun Mo, and Leonidas~J Guibas.
\newblock Pointnet: Deep learning on point sets for 3d classification and segmentation.
\newblock In {\em Proceedings of the IEEE conference on computer vision and pattern recognition}, pages 652--660, 2017.

\bibitem{yan2016perspective}
Xinchen Yan, Jimei Yang, Ersin Yumer, Yijie Guo, and Honglak Lee.
\newblock Perspective transformer nets: Learning single-view 3d object reconstruction without 3d supervision.
\newblock {\em Advances in neural information processing systems}, 29, 2016.

\bibitem{chen2019unpaired}
Xuelin Chen, Baoquan Chen, and Niloy~J Mitra.
\newblock Unpaired point cloud completion on real scans using adversarial training.
\newblock {\em arXiv preprint arXiv:1904.00069}, 2019.

\bibitem{tchapmi2019topnet}
Lyne~P Tchapmi, Vineet Kosaraju, Hamid Rezatofighi, Ian Reid, and Silvio Savarese.
\newblock Topnet: Structural point cloud decoder.
\newblock In {\em Proceedings of the IEEE/CVF Conference on Computer Vision and Pattern Recognition}, pages 383--392, 2019.

\bibitem{mandikal20183d}
Priyanka Mandikal, KL~Navaneet, Mayank Agarwal, and R~Venkatesh Babu.
\newblock 3d-lmnet: Latent embedding matching for accurate and diverse 3d point cloud reconstruction from a single image.
\newblock {\em arXiv preprint arXiv:1807.07796}, 2018.

\bibitem{chen2019text2shape}
Kevin Chen, Christopher~B Choy, Manolis Savva, Angel~X Chang, Thomas Funkhouser, and Silvio Savarese.
\newblock Text2shape: Generating shapes from natural language by learning joint embeddings.
\newblock In {\em Computer Vision--ACCV 2018: 14th Asian Conference on Computer Vision, Perth, Australia, December 2--6, 2018, Revised Selected Papers, Part III 14}, pages 100--116. Springer, 2019.

\bibitem{fu2022shapecrafter}
Rao Fu, Xiao Zhan, Yiwen Chen, Daniel Ritchie, and Srinath Sridhar.
\newblock Shapecrafter: A recursive text-conditioned 3d shape generation model.
\newblock {\em Advances in Neural Information Processing Systems}, 35:8882--8895, 2022.

\bibitem{kim2021conditional}
Jaehyeon Kim, Jungil Kong, and Juhee Son.
\newblock Conditional variational autoencoder with adversarial learning for end-to-end text-to-speech.
\newblock In {\em International Conference on Machine Learning}, pages 5530--5540. PMLR, 2021.

\bibitem{yonekura2021data}
Kazuo Yonekura and Katsuyuki Suzuki.
\newblock Data-driven design exploration method using conditional variational autoencoder for airfoil design.
\newblock {\em Structural and Multidisciplinary Optimization}, 64(2):613--624, 2021.

\bibitem{asperti2020balancing}
Andrea Asperti and Matteo Trentin.
\newblock Balancing reconstruction error and kullback-leibler divergence in variational autoencoders.
\newblock {\em IEEE Access}, 8:199440--199448, 2020.

\bibitem{Nobari.2021}
Amin Heyrani~Nobari, Wei Chen, and Faez Ahmed.
\newblock Range-gan: Range-constrained generative adversarial network for conditioned design synthesis.
\newblock In {\em International Design Engineering Technical Conferences and Computers and Information in Engineering Conference}, volume 85390, page V03BT03A039. American Society of Mechanical Engineers, 2021.

\bibitem{Zhang.2019}
Wentai Zhang, Zhangsihao Yang, Haoliang Jiang, Suyash Nigam, Soji Yamakawa, Tomotake Furuhata, Kenji Shimada, and Levent~Burak Kara.
\newblock 3d shape synthesis for conceptual design and optimization using variational autoencoders.
\newblock In {\em International Design Engineering Technical Conferences and Computers and Information in Engineering Conference}, volume 59186, page V02AT03A017. American Society of Mechanical Engineers, 2019.

\bibitem{Nobariv.2021}
Amin~Heyrani Nobari, Wei Chen, and Faez Ahmed.
\newblock Pcdgan: A continuous conditional diverse generative adversarial network for inverse design.
\newblock {\em arXiv preprint arXiv:2106.03620}, 2021.

\bibitem{Chen.2021b}
Wei Chen and Arun Ramamurthy.
\newblock Deep generative model for efficient 3d airfoil parameterization and generation.
\newblock In {\em AIAA Scitech 2021 Forum}, page 1690, 2021.

\bibitem{qian2020accelerating}
Chao Qian and Wenjing Ye.
\newblock Accelerating gradient-based topology optimization design with dual-model neural networks.
\newblock {\em arXiv preprint arXiv:2009.06245}, 2020.

\bibitem{uugur2019multi}
Mesut U{\u{g}}ur and Ozan Keysan.
\newblock Multi-physics design optimisation of a gan-based integrated modular motor drive system.
\newblock {\em The Journal of Engineering}, 2019(17):3900--3905, 2019.

\bibitem{bae2018perlin}
Hyun-Jin Bae, Chang-Wook Kim, Namju Kim, BeomHee Park, Namkug Kim, Joon~Beom Seo, and Sang~Min Lee.
\newblock A perlin noise-based augmentation strategy for deep learning with small data samples of hrct images.
\newblock {\em Scientific reports}, 8(1):1--7, 2018.

\bibitem{petroll2021synthetic}
Christoph Petroll, Martin Denk, Jens Holtmannsp{\"o}tter, Kristin Paetzold, Philipp H{\"o}fer, et~al.
\newblock Synthetic data generation for deep learning models.
\newblock In {\em DS 111: Proceedings of the 32nd Symposium Design for X (DFX2021)}, pages 1--10, 2021.

\bibitem{madenci2015finite}
Erdogan Madenci and Ibrahim Guven.
\newblock {\em The finite element method and applications in engineering using ANSYS{\textregistered}}.
\newblock Springer, 2015.

\bibitem{kar20223d}
O{\u{g}}uzhan~Fatih Kar, Teresa Yeo, Andrei Atanov, and Amir Zamir.
\newblock 3d common corruptions and data augmentation.
\newblock In {\em Proceedings of the IEEE/CVF Conference on Computer Vision and Pattern Recognition}, pages 18963--18974, 2022.

\bibitem{kingma2014adam}
Diederik~P Kingma and Jimmy Ba.
\newblock Adam: A method for stochastic optimization.
\newblock {\em arXiv preprint arXiv:1412.6980}, 2014.

\bibitem{lorensen1998marching}
William~E Lorensen and Harvey~E Cline.
\newblock Marching cubes: A high resolution 3d surface construction algorithm.
\newblock In {\em Seminal graphics: pioneering efforts that shaped the field}, pages 347--353. 1998.

\bibitem{na2018toxic}
Jonggeol Na, Kyeongwoo Jeon, and Won~Bo Lee.
\newblock Toxic gas release modeling for real-time analysis using variational autoencoder with convolutional neural networks.
\newblock {\em Chemical Engineering Science}, 181:68--78, 2018.

\bibitem{canchumuni2019towards}
Smith~WA Canchumuni, Alexandre~A Emerick, and Marco Aur{\'e}lio~C Pacheco.
\newblock Towards a robust parameterization for conditioning facies models using deep variational autoencoders and ensemble smoother.
\newblock {\em Computers \& Geosciences}, 128:87--102, 2019.

\end{thebibliography}
%%% and comment out the ``thebibliography'' section.

\end{document}